\documentclass[10pt,twocolumn,letterpaper]{article}

\usepackage[pagenumbers]{cvpr} % To force page numbers, e.g. for an arXiv version

\usepackage{graphicx}
\usepackage{amsmath}
\usepackage{amssymb}
\usepackage{booktabs}
\usepackage[linesnumbered,ruled]{algorithm2e}
\newcommand{\name}[0]{LOMA\xspace}
\newcommand{\nameI}[0]{LOMA\_I\xspace}
\newcommand{\nameF}[0]{LOMA\_F\xspace}
\newcommand{\nameIF}[0]{LOMA\_IF\xspace}
\newcommand{\nameAll}[0]{LOMA\_IF\&FO\xspace}

\usepackage{multirow}
\usepackage[pagebackref,breaklinks,colorlinks]{hyperref}

\usepackage[capitalize]{cleveref}
\crefname{section}{Sec.}{Secs.}
\Crefname{section}{Section}{Sections}
\Crefname{table}{Table}{Tables}
\crefname{table}{Tab.}{Tabs.}

\def\cvprPaperID{XXXX} % *** Enter the CVPR Paper ID here
\def\confName{XXXX}
\def\confYear{2023}

\begin{document}

\title{Local Magnification for Data and Feature Augmentation}

\author{Kun He\thanks{The first two authors contributed equally. Correspondence to Kun He.} ~\&  Chang Liu \\
School of Computer Science and Technology\\
Wuhan, China\\
{\tt\small \{brooklet60,liuchanghust\}@hust.edu.cn}
\and
Stephen Lin\\
Microsoft Research Asia\\
\\
{\tt\small stevelin@microsoft.com}
\and 
John E. Hopcroft\\
Department of Computer Science, Cornell University\\
{\tt\small jeh@cs.cornell.edu}
}
\maketitle

%%%%%%%%% ABSTRACT
\begin{abstract}
In recent years, many data augmentation techniques have been proposed to increase the diversity of input data and reduce the risk of overfitting on deep neural networks. In this work, we propose an easy-to-implement and model-free data augmentation method called \textbf{Lo}cal \textbf{Ma}gnification (\name). Different from other geometric data augmentation methods that perform global transformations on images, \name generates additional training data by randomly magnifying a local area of the image. This local magnification results in geometric changes that significantly broaden the range of augmentations while maintaining the recognizability of objects. 
%can increase the amount of training data and destroy the image features by randomly magnifying a specific portion of the image during the training stage. 
Moreover, we extend the idea of \name and random cropping to the feature space to augment the feature map, which further boosts the classification accuracy considerably. 
%Experiments show that our proposed method, though straightforward, can significantly improve baselines for image classification and object detection. The results of adding \name upon the current standard data augmentations, namely random cropping plus random flipping, indicate that our method could be added to the current standard data augmentation suite for higher performance.
Experiments show that our proposed \name, though straightforward, 
can be combined with standard data augmentation to significantly improve the performance on image classification and object detection. And further combination with our feature augmentation techniques, termed \nameAll, can continue to strengthen the model and outperform advanced intensity transformation methods for data augmentation. 
%The models and code will be made publicly available.
\end{abstract}
%standard data augmentation (random cropping and random flipping)
% resize and random flipping 

%%%%%%%%% BODY TEXT
\section{Introduction}
\label{sec:intro}
Convolutional neural networks (CNNs) have achieved broad success over a variety of computer vision tasks, including image classification~\cite{randomcrop,randomflip,Vit}, object detection~\cite{fastrcnn,yolo}, and semantic segmentation~\cite{seg2,seg}. However, overfitting often occurs during training and reduces model performance. %To increase the valid training data and improve the generalization of neural networks, 
A widely used technique to combat this problem is data augmentation, which expands the set of valid training data and improves the generalization of neural networks.

\begin{figure*}[ht]
  \centering
%   \fbox{\rule{0pt}{2in} \rule{0.9\linewidth}{0pt}}
  \includegraphics[width=0.8\linewidth]{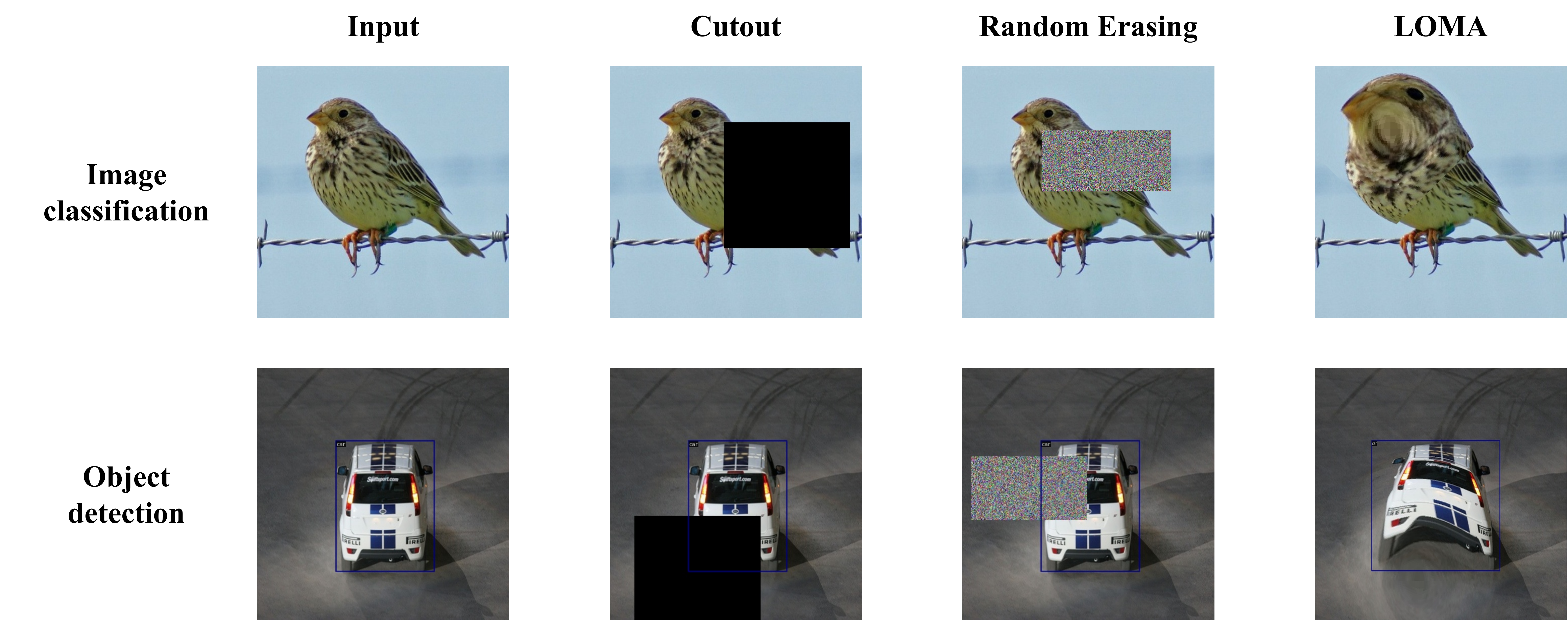}
  %\text{~~~~~~~~~~~~~~~~~~~~~~~~~~~~~~~~~~~~~~~~~~~~~~~~~~~~~~~~~~Input~~~~~~~~~~~~~~~~~~~~~~~~ Cutout~~~~~~~~~~~~~~~~~Random Erasing ~~~~~~~~~~~~~~LOMA~~~~~~~~~~~~~~~~~~~~~~~~~~}
   \caption{Illustration of augmented images by various methods, including Cutout, Random Erasing and our Local Magnification (\name).}
   \label{fig1}
\end{figure*}

Mainstream image augmentation methods can be split into three categories~\cite{xu2022comprehensive}, namely model-free, model-based, and optimization-based. Among them, model-free methods such as Cutout~\cite{cutout}, Hide and Seek~\cite{singh2018hide}, Random Erasing~\cite{randomerase}, and GridMask~\cite{chen2020gridmask} have been especially popular due to their relative simplicity and effectiveness. As illustrated in~\cref{fig1}, model-free data augmentations mainly focus on occluding some of the image content via intensity transformations so that the CNNs can be pushed to learn more robust features. However, when an essential portion of an image is occluded, the image label may become ambiguous. For instance, when the data augmentation completely masks valuable objects in the input image, the image data becomes noisy and harmful to the training of CNNs.

To avoid such label ambiguity in model-free data augmentation, 
%To avoid changing the label when modifying the input images and using the augmented images as the training data to improve the model generalization, 
we introduce a new data augmentation method, termed \textbf{Lo}cal \textbf{Ma}gnification (\name). Unlike other geometric augmentations that are based on global geometric transformations, such as affine transformations, elastic distortions~\cite{ElasticDistortion}, random cropping~\cite{randomcrop}, and random flipping~\cite{randomflip}, \name only magnifies a local portion of the image to generate the augmentation, making some local features more prominent. 

In the physical world, occlusion can happen where some portion of a scene is occluded by other objects, while our \name can also occur like when a person uses a magnifier on an input image to facilitate focusing on some details. Another case is that cars can be severely deformed in a traffic accident yet still be well recognized. Also, human faces can be exaggeratedly distorted in funhouse curved mirrors but still be recognized from the distorted images by people.

\begin{figure}[t]
\centering
%\framebox[4.0in]{$\;$}
% \includegraphics[width=10.5cm]{fig/fig2.pdf}
\includegraphics[width=0.9\linewidth]{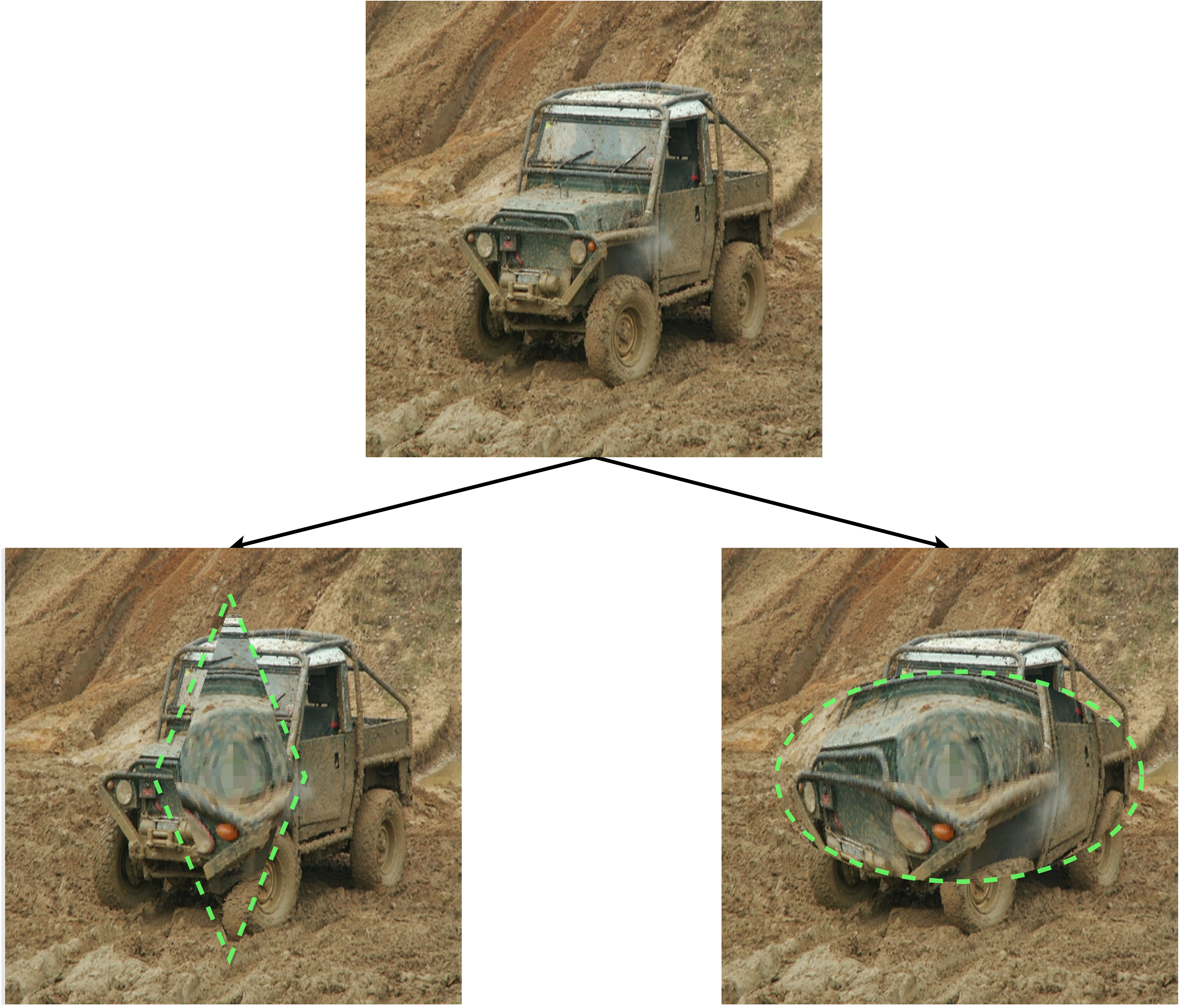}
\caption{Examples of \name for two preset deformation shapes (rhombus and ellipse).}
\label{fig2}
\end{figure}

In network training, \name is applied to each image of a mini-batch with a fixed probability. First, a magnification center is randomly located on each selected image, and a radius is determined within a specified range. Then, \name calculates new horizontal and vertical coordinates for all the pixels located within a neighborhood of the magnification center that is defined by the radius and a preset deformation shape. These pixels are moved accordingly to obtain the new image, as shown for two preset deformation shapes (rhombus and ellipse) in~\cref{fig2}. Although this procedure may discard a small portion of the pixels, the number of discarded pixels is much less than that of intensity transformation methods based on occlusion. In addition, some augmented images will have apparent dislocations at the boundaries of the augmented region, which cannot be achieved by intensity transformation. 
We argue that these deformations and dislocations increase the diversity of the training data while maintaining the apparent image labels, providing challenging yet effective training data that goes well beyond the standard augmentations.

% Moreover, we extend the idea of our local magnification as well as the standard random cropping to augment feature maps in the feature space. The former is called \name on the Feature Map, and the latter is called FeatureMap Offset. We abbreviate the two methods used together as \nameIF (\textbf{\name} operating in the \textbf{F}eature space). For FeatureMap Offset, it offsets the feature map output at an intermediate layer of the network with a certain probability and then continues as an input of the subsequent layer.
We additionally extend the idea of our local magnification as well as the standard random cropping to augment feature maps in the feature space. The former is called \name on the Feature Map, and the latter is called FeatureMap Offset (FO). We abbreviate the joint use of \name and \name on the Feature Map as \nameIF (\textbf{\name} in the \textbf{I}mage space and \textbf{F}eature space). For FeatureMap Offset, it offsets the feature map output at an intermediate layer of the network with a certain probability and then passes it to the subsequent layer.

Our main contributions can be summarized as follows:
\begin{itemize}
\item We propose a model-free data augmentation method based on local geometric transformations called Local Magnification (\name). It is a lightweight method that does not require additional parameters to learn and can be easily plugged into the training of various CNNs.
\item The proposed \name improves the baseline (standard augmentation) for image classification on the CIFAR and ImageNet benchmark datasets and for object detection on the PASCAL VOC benchmark dataset. \name can be combined with standard data augmentation to further boost the performance.
\item  We extend the idea of \name and random cropping to the feature space. %, termed \nameF and FO. %The results indicate that using \name both in image space and feature space, termed \nameIF still performs well.
% We extend the idea of \name and random cropping to the feature space, termed \nameF and FO. The results indicate that our methods still performs well. 
%and can be incorporated with \name on the input image to further boost performance.
%\item 
The incorporation of our final method, termed \nameAll, can further boost performance and outperform two advanced intensity transformation data augmentation methods, Cutout~\cite{cutout} and Random Erasing~\cite{randomerase}.
% Here each method is superimposed on standard data argumentation.
\end{itemize}

%-------------------------------------------------------------------------
\section{Related Work}
\label{sec:RW}
\subsection{Data Augmentation}
In recent years, many data augmentation techniques have been proposed to address the overfitting issue that arises when training convolutional neural networks (CNNs)~\cite{xu2022comprehensive}. Existing data augmentation methods can be categorized into three groups, namely model-free, model-based, and optimizing policy-based.

Model-based approaches attempt to augment the training dataset with images produced by Generative Networks. Antoniou \etal~\cite{DAGAN} use a generative model called DCGAN to take data from a source domain and generalize it to generate other within-class data. Huang \etal~\cite{huang2018auggan} propose a structure-aware image-to-image translation network called AugGAN to learn the joint distribution of the two domains and find transformations between them. Xu \etal~\cite{xu2022masked} adopt the self-supervised masked autoencoder to generate distorted views of the input images as augmented images.

Optimizing policy-based image augmentation strategies seek to find the optimal combination policy of multiple data augmentation methods through reinforcement learning or adversarial learning. The most representative work is AutoAugment~\cite{autoaugment}, which has demonstrated powerful performance on various benchmarks but spends much time searching for policies. There are many follow-up improvements, such as Fast AutoAugment~\cite{lim2019fast}, Adversarial AutoAugment~\cite{zhang2019adversarial}, RandAugment~\cite{randaugment}, Faster AutoAugment~\cite{hataya2020faster}, and TeachAugment~\cite{suzuki2022teachaugment}.

Most relevant to our work are model-free data augmentation methods. In addition to methods that process multiple images to obtain new samples, such as Mixup~\cite{mixup}, CutMix~\cite{cutmix}, and PuzzleMix~\cite{puzzlemix}, there are single-image augmentation strategies, which can be divided into three branches: geometric transformation, color image processing, and intensity transformation. Color image processing methods perturb the color of the input image by transforming the color of pixels within the training data. In the intensity transformation methods, DeVries and Taylor~\cite{cutout} propose Cutout, which uses a square fixed-size mask to randomly occlude a portion of the image and fill it with uniform pixels. Zhong \etal~\cite{randomerase} propose Random Erasing, which uses random pixel values to form a rectangular mask. To establish a good balance between deletion and the preservation of regional information on the images, Chen \etal~\cite{chen2020gridmask} introduce GridMask, which randomly removes non-adjacent image patches. 
The above methods all perform some occlusion so as to encourage the network to focus more on complementary and less prominent features~\cite{cutout}. 
%In contrast, our method alters the position of pixels to magnify a local area of the image, rather than block a portion of the image directly.
Instead of blocking a portion of the image directly, our method alters the position of pixels to magnify a local area of the image, making it more prominent and helping the network to focus more on these local features. As the local region is located randomly, the network can learn to generalize to more extensive variations of the original image. 

\subsection{Regularization}
Another effective approach to address the overfitting issue in CNNs is by regularization in the feature space. Dropout~\cite{dropout} randomly drops units and their connections from the neural network during training. Follow-ups to Dropout contain many improvements, such as DropBlock~\cite{dropblock} which occludes continuous regions on the feature map. Another idea is to exchange some components of the two feature maps. For example, PatchUp~\cite{patchup} mixes consecutive feature blocks in the feature space. MoEx~\cite{moex} forces the model to improve robustness by randomly replacing the moments of the feature map of one training sample with those of another and interpolating the labels. 
These methods can also be regarded as augmentations in the feature space. 
Thereby, we apply our \name in the feature space as another form of regularization, and we further consider porting standard data augmentations, namely random cropping and random flipping, to the feature space as well. As random flipping in the image space and feature space have similar effects, we only adopt random cropping in the feature space and refer to this new method as \textbf{F}eatureMap \textbf{O}ffset (FO). 
%Our method can also be applied on the feature space and achieve good results.

\subsection{Standard Data Augmentation Suite}
The two most successful data augmentation techniques for training CNNs are the geometric transformations of random cropping~\cite{randomcrop} and random flipping~\cite{randomflip}. Through these two types of augmentations, the diversity of training samples is increased. Unlike random cropping and random flipping which perform a global transformation of an image, \name can be regarded as a local %noise
geometric transformation because it does not alter the overall structure of the image. Our method can be combined with random cropping and random flipping as a new data augmentation baseline for training CNNs.

%-------------------------------------------------------------------------
\section{Methodology}
\label{sec:03Method}
In this section, we introduce the proposed LOMA in detail. \Cref{LOMA in the image space} explains how LOMA functions at the input layer. \Cref{LOMA in the feature space} describes how to extend the idea of \name and random cropping to the feature space.

\subsection{LOMA in the Image Space}
\label{LOMA in the image space}
\begin{figure}[t]
  \centering
  \begin{subfigure}{0.49\columnwidth}
    \centering
    \includegraphics[width=0.99\columnwidth]{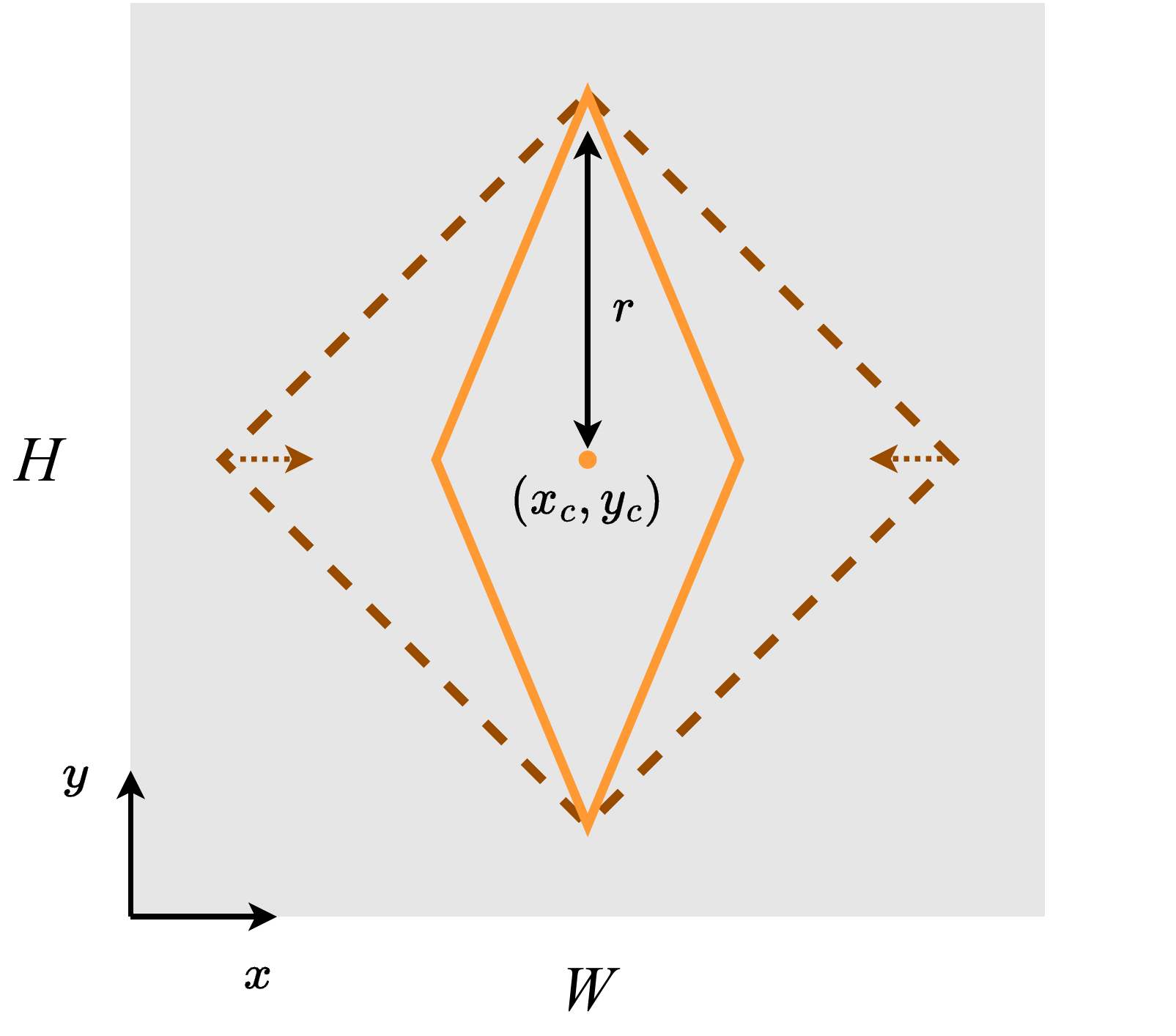}
    \caption{Rhombus}
    \label{fig3:subfig:a}
  \end{subfigure}
  \hfill
  \begin{subfigure}{0.49\columnwidth}
    \centering
    \includegraphics[width=0.99\columnwidth]{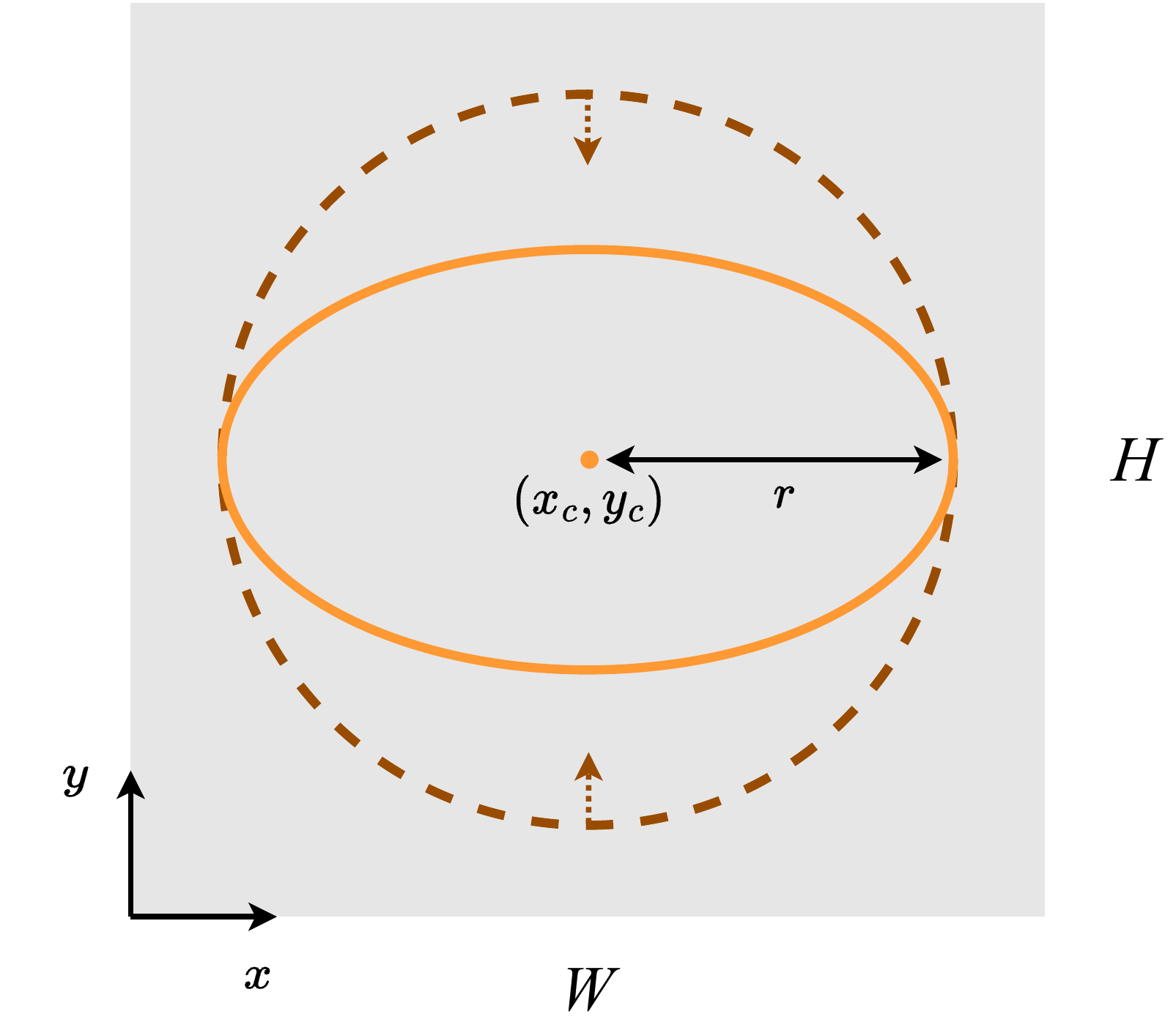}
    \caption{Ellipse}
    \label{fig3:subfig:b}
  \end{subfigure}
  \caption{The area enclosed by the dashed brown line is the preset deformation shape, and the area enclosed by the solid orange line is the actual deformation shape. }
  \label{fig3}
\end{figure}

Local Magnification (\name) is an easy-to-implement method that augments the training dataset effectively. In the training phase, for each input image in a mini-batch, our algorithm is randomly applied on the image with a certain probability $p$. Thus, the model receives the augmented image with probability $p$, and the probability of receiving the unaltered image is $1-p$.  We use a vector $(x_c,y_c,r,s,a_x,a_y)$ to describe the region $S$ where \name deforms an image $I$. As shown in~\cref{fig3} for an input image size of $W \times H$, we create a coordinate frame with the bottom left corner of the image as the origin. \name randomly determines a magnification center on the image, denoted as $(x_c, y_c)$.
%where we denote the abscissa of the magnification center as $x_c$ and the ordinate as $y_c$. 
$s$ is the preset deformation shape, 
% (we choose rhombus or ellipse in this paper, but other shapes like rectangle could also work)
and $r$ is the radius. $a_x$, $a_y$ are the horizontal and vertical compression ratios relative to the preset deformation shape.

% \begin{figure}[t]
% \centering
% \includegraphics[width=0.95\linewidth]{fig/fig3.pdf}\\

% \text{ (a) Ellipse~~~~~~~~~~~~~~~~~~~~~~~~~ (b) Rhombus}

% \caption{The area enclosed by the dashed brown line is the preset deformation shape, and the area enclosed by the solid orange line is the actual deformation shape.}
% \label{fig3}
% \end{figure}

For the preset deformation shapes, we choose to define them using the $\ell_1$ and $\ell_2$ norm (rhombus and ellipse) in this work to facilitate calculation. Note that other shapes like a rectangle, which corresponds to the $\ell_{\infty}$ norm, could also work. 

For the $\ell_1$ norm, the preset deformation shape $s$ is a rhombus, and the deformation region $S$ is defined as:
\begin{equation}
    \{(x_i,y_i) \big|~~ a_x\left| x_i-x_c \right| + a_y \left| y_i-y_c \right| < r\},
    \label{equ:1}
\end{equation}
where $(x_i, y_i)$ is the %abscissa and ordinate 
coordinates of each pixel inside the deformation region. 

For the $\ell_2$ norm, the preset deformation shape $s$ is an ellipse, and the deformation region $S$ is defined as:
\begin{equation}
     %\{(x_i,y_i)\big|~~ \sqrt{ a_x( x_i-x_c )^{2} + a_y ( y_i-y_c )^{2} } < r}\}.
     \{(x_i,y_i)\big|~~ a_x( x_i-x_c )^{2} + a_y ( y_i-y_c )^{2} < r^{2} \}.
    \label{equ:2}
\end{equation}

For each pixel $(x_i, y_i)$ in $S$, \name calculates a new position of this pixel and moves it accordingly to obtain the output image $I^{+}$:
\begin{equation}
\left\{
        \begin{array}{lr}
         x_o = \frac{d}{r} \cdot (x_i - x_c) + x_c, \\
         y_o = \frac{d}{r} \cdot (y_i - y_c) + y_c, \\
         I^{+} [x_i,y_i] \leftarrow I[x_o,y_o] \\
        \end{array}
        %  \label{equ:3}
\right.
\label{equ:3}
\end{equation}
where $d = \sqrt{(x_i-x_c)^{2}+(y_i-y_c)^{2}}$ is the distance between the pixel to be changed and the magnification center. For each input image, $r$ is randomly set within a given range as
% $r = random(r_{min} \times max(H,W),r_{max} \times max(H,W))$
$r = random(r_{min},r_{max}$) $\times \; max(H,W)$, and either $a_x$ or $a_y$ takes a random value within the given range $random(a_{min},a_{max})$ while the other is set to $1$. $r$ determines the scope of local magnification in the image, while $a_x$ and $a_y$ determine the intensity of the deformation within the preset deformation region. 

The overall process is shown in~\cref{alg:PM}. We denote this method as \nameI and also refer to it as \name for brevity. % which is also consistent with other data augmentation methods. 

\begin{algorithm}[t]
\caption{Local Magnification on Image}
\label{alg:PM}
\DontPrintSemicolon
    \SetAlgoLined
    \KwIn {input image $I$, image size $(W, H)$, probability $p$, preset deformation shape $s$, 
    horizontal/vertical compression ratio range $(a_{min}, a_{max})$, radius range $(r_{min}, r_{max})$}
    \KwOut {augmented image $I^{+}$}
    % $p_r \gets random(0,1)$\;
    % $p_c \gets random(0,1)$\;
    $I^+ \gets I$\;
    \eIf {$random(0,1)>p$}{
    return $I^+$\;
    }{
    %$r \gets random(r_{min} \times max(H,W),r_{max} \times max(H,W)$)\;
    $r \gets random(r_{min},r_{max}$) $\times max(H,W)$ 
    
        \eIf{ $random(0,1)>0.5$ }{
        $a_x \gets random(a_{min},a_{max})$;
        $a_y \gets 1$
        }{
        $a_x \gets 1$;
        $a_y \gets random(a_{min},a_{max})$
        }
    
    \eIf{$s = $ `rhombus'}{
       %$ S \gets \{ (x_i,y_i)| a_x\left| x_i-x_c \right| + a_y \left| y_i-y_c \right| <r\}$\;
       Calculate $S$ by Eq. (\ref{equ:1})%Eq. (1);
    }{
       % $S \gets \{ (x_i,y_i)| a_x( x_i-x_c )^{2} + a_y ( y_i-y_c )^{2} <r^{2}\}$\;
       Calculate $S$ by Eq. (\ref{equ:2})%Eq. (2);
    }
    \For{$(x_i,y_i)\in S$}{
        $d \gets \sqrt{(x_i-x_c)^{2}+(y_i-y_c)^{2}}$\;
        %$x_o \gets \frac{d}{r} * (x_i – x_c) + x_c$\;
        %$y_o \gets \frac{d}{r} * (y_i – y_c) + y_c$\;
        %$I^{+} [x_i,y_i] \gets I[x_o,y_o]$\;
        Calculate $x_o, y_o, $ and $I^{+}[x_i,y_i]$ by Eq. (\ref{equ:3}) %Eq. (3);
    }
    return $I^+$\;
    }
\end{algorithm}

\subsection{LOMA in the Feature Space}
\label{LOMA in the feature space}
Unlike the above augmentation in the image space, that applies \name with probability $p$ to each image in the mini-batch, \name in the feature space augments each mini-batch with probability $p_f$ in an intermediate layer of the neural network. Since adapting the idea of \name to an intermediate layer is straightforward, we focus on how to extend the idea of random cropping into the feature space, which we call FeatureMap Offset (FO). 

\begin{figure}[t]
\centering
\includegraphics[width=0.9\linewidth]{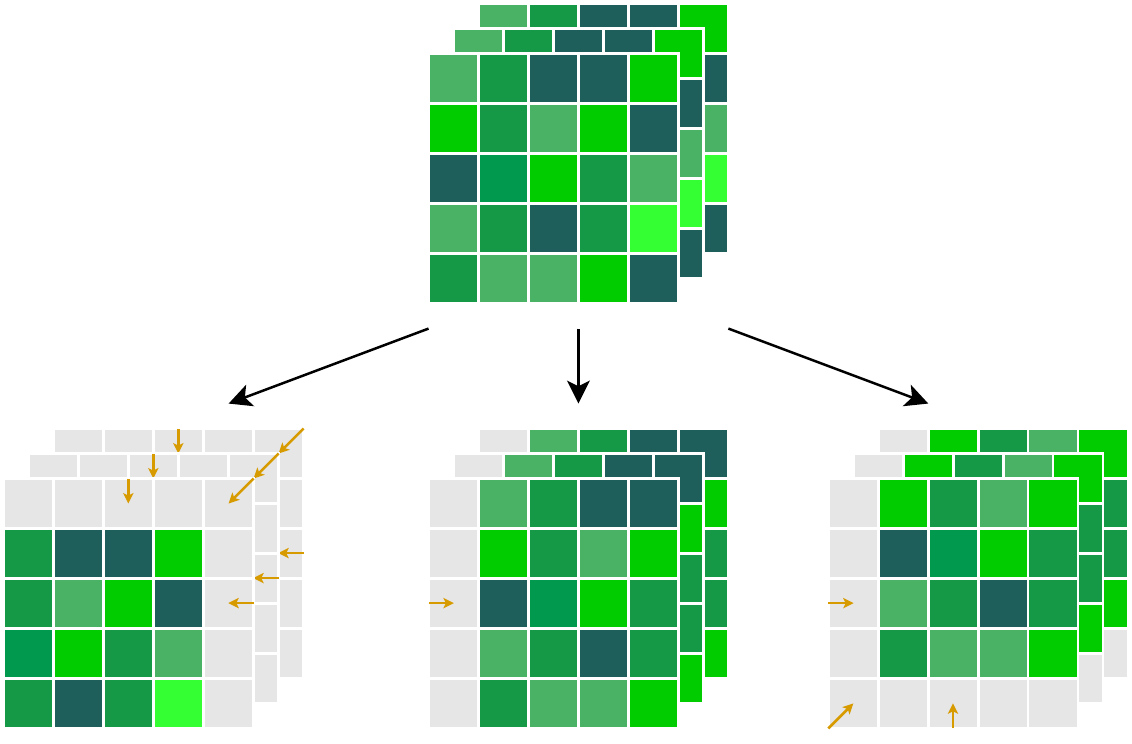}

%\caption{The FeatureMap Offset on the feature map. }
\caption{Examples of offsets in FeatureMap Offset (FO) method. }
\label{fig4}
\end{figure}

% For convenience, the preset deformation shape of \name on the Feature Map is set to rhombus in the experiments. 
We decide with probability $p_f$ whether to use both \name on the Feature Map and FeatureMap Offset simultaneously for a mini-batch of feature maps that comes from the $l^{th}$ convolutional block of the network (a smaller $l$ is closer to the input). 
But note that the two methods can also be used separately. 
FeatureMap Offset does not change the size of the feature map but applies the translation operation of an affine transformation to move the feature map and fill the vacated entries with 0. Three possible scenarios are shown in \cref{fig4}, with the feature maps shifted to the lower left, right and upper right, respectively.

For each input feature map, the affine transformation matrix of FeatureMap Offset is:
\begin{equation}
  M = \left[
  \begin{array}{ccc}
  1 & 0 & T_x\\
  0 & 1 & T_y\\
  0 & 0 & 1\\
  \end{array}
  \right], 
  \label{equ:4}
\end{equation}
where $T_x$ and $T_y$ represent the offset of the feature map in the horizontal and vertical directions. Assuming the feature map size to be $B \times C \times H_f \times W_f$, then $T_x$ and $T_y$ each takes a random value in the given range
% $random(-\gamma \times max(H_f,W_f),\gamma \times max(H_f,W_f))$
$random(-\gamma ,\gamma)\times max(H_f,W_f)$
, where $\gamma$ controls the offset degree of the feature map.

%We abbreviate the two methods used together as \nameIF (\textbf{\name} in \textbf{F}eature space) for description in the experimental section.
%We abbreviate the two methods used together as LOMA\_F\&FO (\textbf{\name} in \textbf{F}eature space and FeatureMap Offset). 
We abbreviate \name on the Feature Map as \nameF, abbreviate \name in the image space and feature space as \nameIF, and denote \nameIF together with FeatureMap Offset as \nameAll.

%-------------------------------------------------------------------------
\section{Experiments}

\begin{table*}[ht]
\centering
\resizebox{\textwidth}{!}{
\begin{tabular}{l|ccc|ccc}
 \hline
%\multicolumn{1}{c|}{} 
\multirow{2}{*}{Method}
& \multicolumn{3}{c|}{CIFAR-10}          & \multicolumn{3}{c}{CIFAR-100}          \\
%\multicolumn{1}{c|}{}
& ResNet-20 & WRN-28-10 & PyramidNet & ResNet-20 & WRN-28-10 & PyramidNet \\ \hline
Baseline              &92.75 $\pm$ 0.26~~    &96.22 $\pm$ 0.10~~     &96.16 $\pm$ 0.18~~ &69.16 $\pm$ 0.28~~     &81.32 $\pm$ 0.20~~ &83.68 $\pm$ 0.34~~ \\
Cutout~\cite{cutout}&  \textbf{93.67} $\pm$ 0.21~~   &96.92 $\pm$ 0.16$^*$ &     96.90$^{**}$~~~~~~~~~~ &  70.16 $\pm$ 0.26~~ &81.59 $\pm$ 0.27$^*$& 83.47$^{**}$~~~~~~~~~~~~\\
Random Erasing~\cite{randomerase}&93.27 $\pm$ 0.09$^*$&96.92 $\pm$ 0.05$^*$ &     -           &70.03 $\pm$ 0.11$^*$ &82.27 $\pm$ 0.15$^*$& -\\ \hline
LOMA (rhombus)      &93.27 $\pm$ 0.10~~    &96.85 $\pm$ 0.13~~     &96.73 $\pm$ 0.32~~ &70.12 $\pm$ 0.28~~     &82.28 $\pm$ 0.23~~ &83.91 $\pm$ 0.48~~ \\
LOMA (ellipse)      &93.26 $\pm$ 0.10~~    &96.89 $\pm$ 0.11~~     &\underline{96.79} $\pm$ 0.25~~ &70.15 $\pm$ 0.19~~     &81.88 $\pm$ 0.21~~ &83.74 $\pm$ 0.19~~ \\
% + LOMA\_F             &92.95 $\pm$ 0.05~~    &                     &                 &69.81 $\pm$ 0.11~~     &                &        \\
LOMA\_IF     &93.42 $\pm$ 0.17~~    &  \textbf{97.05} $\pm$ 0.12~~  & 96.74 $\pm$ 0.43~~  &\underline{70.34} $\pm$ 0.36~~     &\underline{82.65} $\pm$ 0.30~~   &\underline{84.29} $\pm$ 0.58~~ \\ 
\nameAll     &\underline{93.53} $\pm$ 0.15~~ &\underline{96.97} $\pm$ 0.13~~ &\textbf{96.91} $\pm$ 0.60~~ & \textbf{70.57} $\pm$ 0.21~~ & \textbf{82.89} $\pm$ 0.23~~ &  \textbf{84.45} $\pm$ 0.28~~          \\  \hline 
%LOMA$^e$\_IF & & & &\\
%LOMA$^e$\_IF\&FO & & & &\\  \hline 
\end{tabular}
}
\caption{Test accuracy (\%) on CIFAR-10 and CIFAR-100. 
%Combining LOMA (rhombus) and LOMA\_F yields the best results. 
$^*$ indicates results reported in the original paper. $^{**}$ indicates results reported in another paper~\cite{cutmix} for comparison. The best results are in \textbf{bold} and the second best results are \underline{underlined}.}
\label{table1}
\end{table*}

In this section, we verify the effectiveness of the proposed \name, \nameF and FO. We first conduct multiple experiments on the image classification task using the CIFAR-10, CIFAR-100~\cite{cifar}, and ImageNet~\cite{imagenet} datasets (\cref{Results on CIFAR-10 and CIFAR100} and \cref{ImageNet}). Next, we study the impact of hyper-parameters on model performance and the robustness of the model to occlusion (\cref{Ablation Study} and \cref{Robustness to occlusion}). Then, we demonstrate the complementarity of \name to standard data augmentation (\cref{Complementary to standard data augmentation}). Finally, we further evaluate our method on the PASCAL VOC 2007 detection benchmark~\cite{pascal} for object detection (\cref{Object Detection}).

\subsection{Results on CIFAR-10 and CIFAR100}
\label{Results on CIFAR-10 and CIFAR100}

CIFAR-10 and CIFAR-100~\cite{cifar} both have 60,000 color images of size 32$\times$32, and each dataset is divided into a training set of 50,000 images and a test set of 10,000 images. There are 10 and 100 different classes in CIFAR-10 and CIFAR-100, respectively. We train ResNet-20~\cite{resnet} and WideResNet-28-10~\cite{wideresnet} for 300 epochs using standard data augmentation (padding to 40$\times$40, random cropping, and random horizontal flipping) and Local Magnification (\name). And we utilize the Stochastic Gradient Descent (SGD) optimizer with Nesterov momentum of 0.9 and weight decay of 5e-4. The learning rate is set to 0.1 and decays by a factor of 10 at 100, 200, and 265 epochs. We also use PyramidNet-200~\cite{pyramidal} with widening factor $\tilde{\alpha}= 240$ in the same way as in Cutmix~\cite{cutmix}, where the batch size and training epochs are set to 64 and 300, respectively. The learning rate is set to 0.25 and decays 10-fold at 150 and 225 epochs. In the following experiments, we set $p=0.5$, $r_{max}=0.7$, $r_{min}=0.03$, $a_{max}=3$, $a_{min}=1$ for the hyper-parameters of \name, and $p_f=0.5$, $l=2$, $\gamma=0.25$ for the hyper-parameters of \nameF and FO, if not specified. 

The results on CIFAR-10 and CIFAR-100 are summarized in \cref{table1}. 
%Significant improvements are found after superimposing \name, \nameF, and FO to the baseline model (with standard data augmentation). 
We see that \name (rhombus) and \name (eclipse) yield similar results. 
To avoid confusion, we only use \name (rhombus) for the following comparisons if not written explicitly.
From the table, we can observe the following:

\begin{itemize}
\setlength{\itemsep}{0pt}
\setlength{\parsep}{0pt}
\setlength{\parskip}{0pt}
\item When augmenting the image using \name at the input layer, our method improves the performance of the baseline significantly. The results of \name are competitive to, though do not surpass, two advanced intensity transformation data augmentation methods, Cutout~\cite{cutout} and Random Erasing~\cite{randomerase}.
%we achieve 0.52\%, 0.63\%, and 0.57\% improvements on ResNet-20, WideResNet-28-10 and PyramidNet-200 compared to the baseline, respectively. On CIFAR-100, we observe 0.96\%, 0.96\%, and 0.23\% respective improvements from the baseline model due to \name. 
\item Using \name both in the image space and feature space, the resulting method \nameIF further boosts the performance, showing results similar to Cutout and Random Erasing on CIFAR-10 and outperforming them on CIFAR-100.  
\item Our final method \nameAll outperforms Cutout and Random Erasing in almost all cases, except on CIFAR-10 using a ResNet-20 model. The superiority of \nameAll is more prominent on more complex models and complicated data.    
%, with all models outperforming the case where \name is used by itself at the input layer. 
%Our method obtains a top-1 test accuracy of 97.05\% on CIFAR-10 and 84.45\% on CIFAR-100 when using WideResNet-28-10 and PyramidNet-200, respectively.
%\name can provide competitive results for most networks compared to the intensity transformation methods if only augmented at the input layer, and the effect of \nameAll outperforms Cutout\cite{cutout} and Random Erasing\cite{randomerase}.
\end{itemize}
%------------------------------------------------------------------------

\subsection{Results on ImageNet}
\label{ImageNet}
The ILSVRC 2012 classification dataset~\cite{imagenet}, ImageNet, is the most challenging dataset for the image classification task. It contains about 1.2 million training images and 50,000 validation images with 1000 classes.

\begin{table}[ht]
\centering
\begin{tabular}{lcc}
\hline
\multirow{2}{*}{Method} & \multicolumn{2}{c}{ResNet-50} \\ \cline{2-3} 
                        & Top-1 Err.    & Top-5 Err.    \\ \hline
Baseline                &    23.00~~~      &  6.67~~~          \\
Cutout~\cite{cutout}    &    22.93$^{**}$     &     6.66$^{**}$      \\
Random Erasing~\cite{randomerase}&  22.75$^*$~~   &  6.69$^*$~~   \\ \hline
\name (rhombus)          &     22.63~~~     &  6.69~~~         \\
\name (ellipse)           &     22.79~~~     &  6.59~~~         \\
% \nameF   &   &\\
\nameIF   & 22.57~~~  & 6.40~~~~\\  
\nameAll        &   \textbf{22.33}~~~     &   \textbf{6.30}~~~        \\ \hline
\end{tabular}
\caption{Test error (\%) of ResNet-50 on the ImageNet dataset (the lower the better). $^*$ indicates results reported in the original paper. $^{**}$ indicates results reported in another paper~\cite{cutmix} for comparison.}
\label{table2}
\end{table}

\begin{figure}[t]
  \centering
  %\vspace{-1em}
%   \fbox{\rule{0pt}{2in} \rule{0.9\linewidth}{0pt}}
  \includegraphics[width=0.95\linewidth]{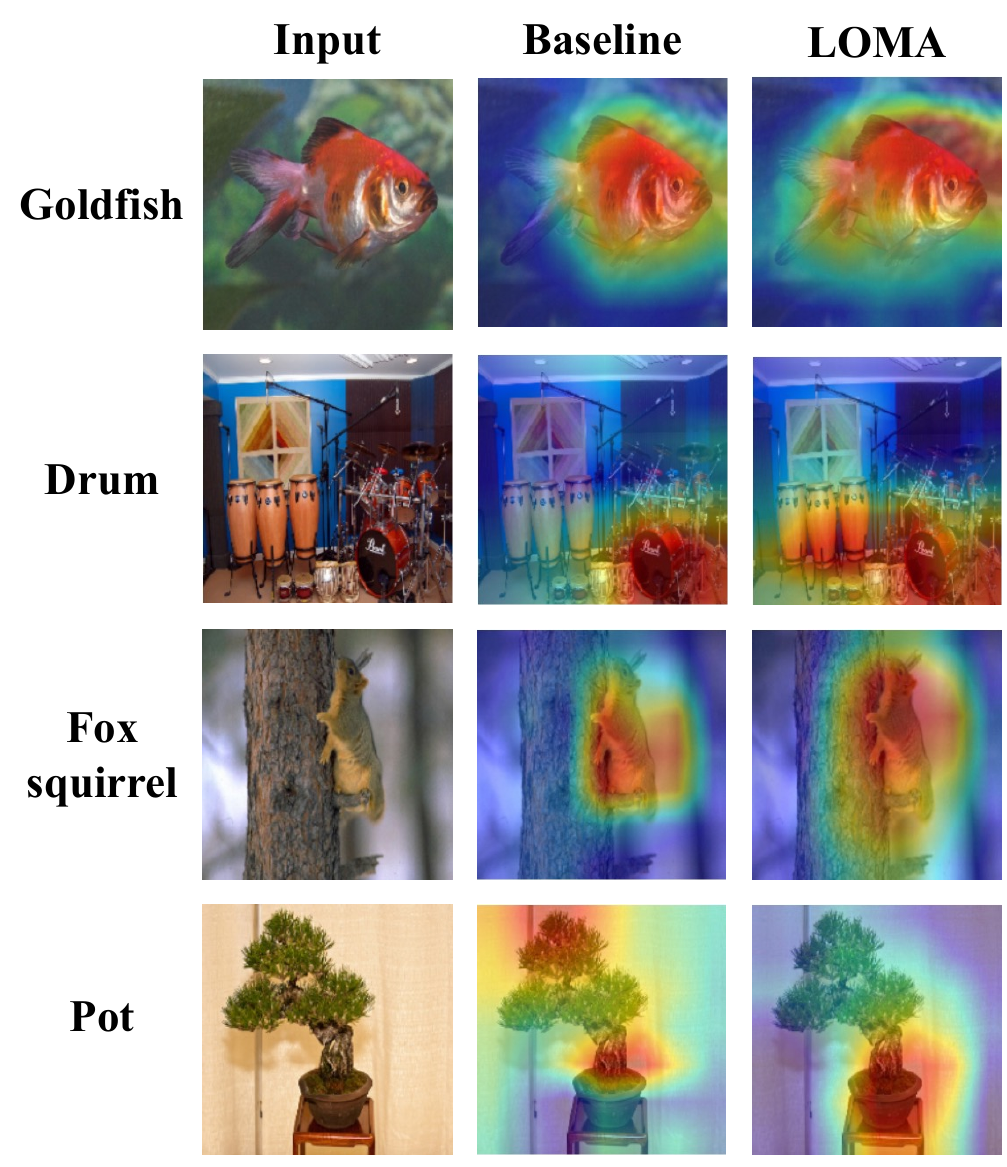}
  \caption{Class activation mapping (CAM)~\cite{CAM} visualizations for ResNet-50 trained on ImageNet. %Compared to the baseline, 
  \name can force the network to focus on a larger range or more important region.}
  \label{fig5}
\end{figure}

\begin{figure*}[ht]
  \centering
  \begin{subfigure}{0.32\linewidth}
    % \fbox{\rule{0pt}{2in} \rule{.9\linewidth}{0pt}}
    \centering
    \includegraphics[width=0.95\linewidth]{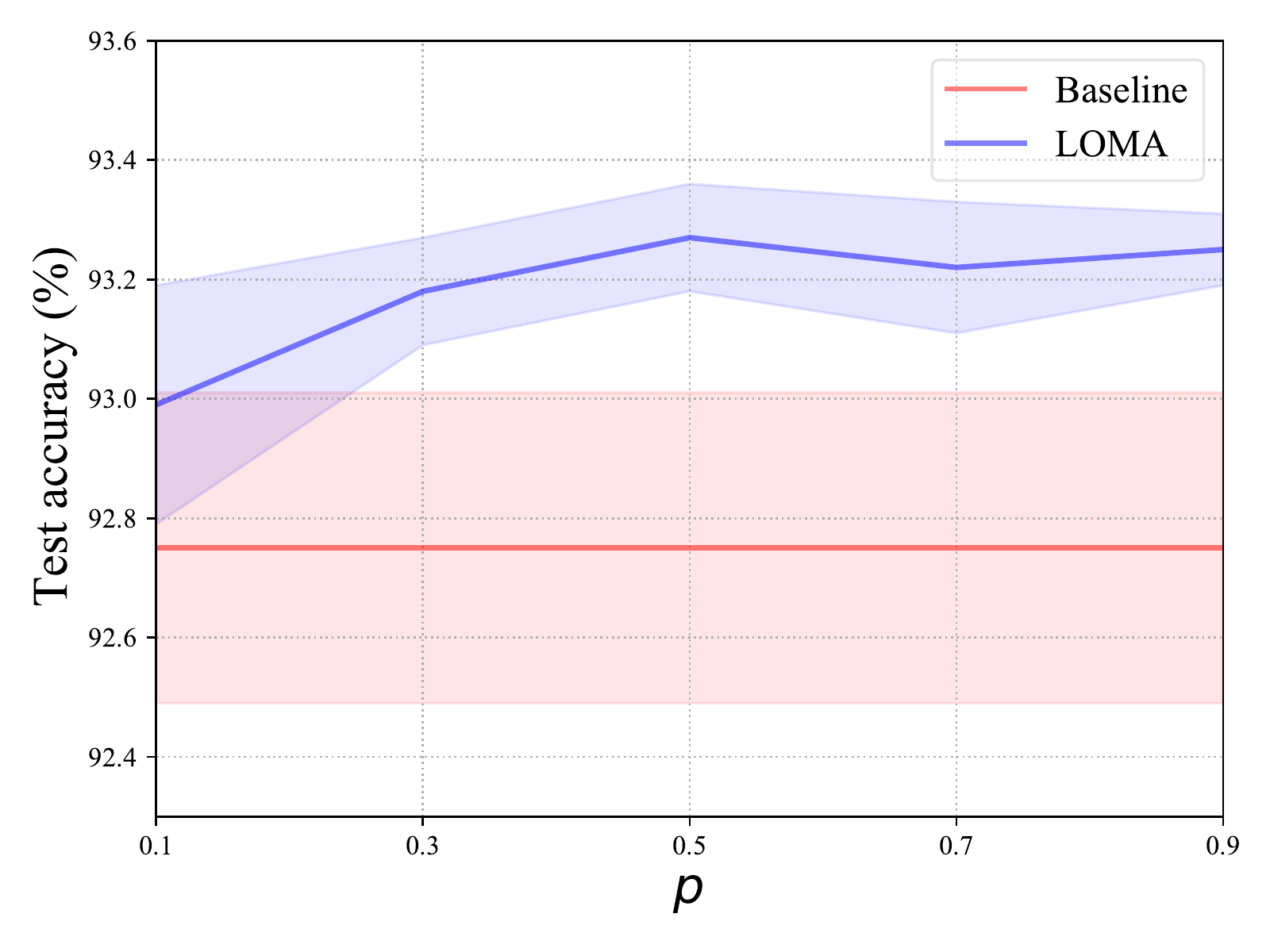}
    \caption{Probability $p$}
    \label{fig:subfig:a}
  \end{subfigure}
  \hfill
  \begin{subfigure}{0.32\linewidth}
    \centering
    % \fbox{\rule{0pt}{2in} \rule{.9\linewidth}{0pt}}
    \includegraphics[width=0.95\linewidth]{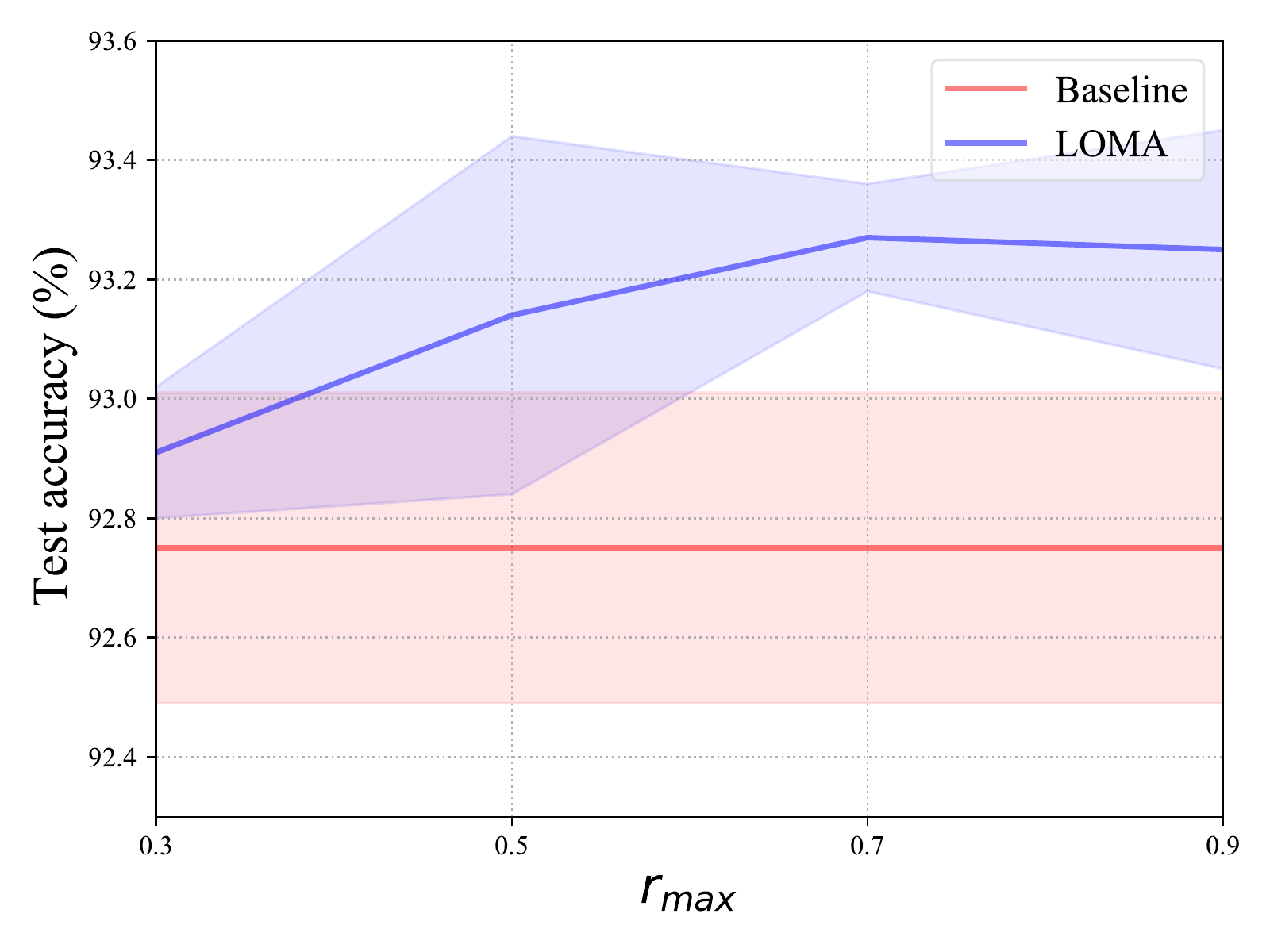}
    \caption{Radius $r_{max}$}
    \label{fig:subfig:b}
  \end{subfigure}
  \hfill
  \begin{subfigure}{0.32\linewidth}
    % \fbox{\rule{0pt}{2in} \rule{.9\linewidth}{0pt}}
    \centering
    \includegraphics[width=0.95\linewidth]{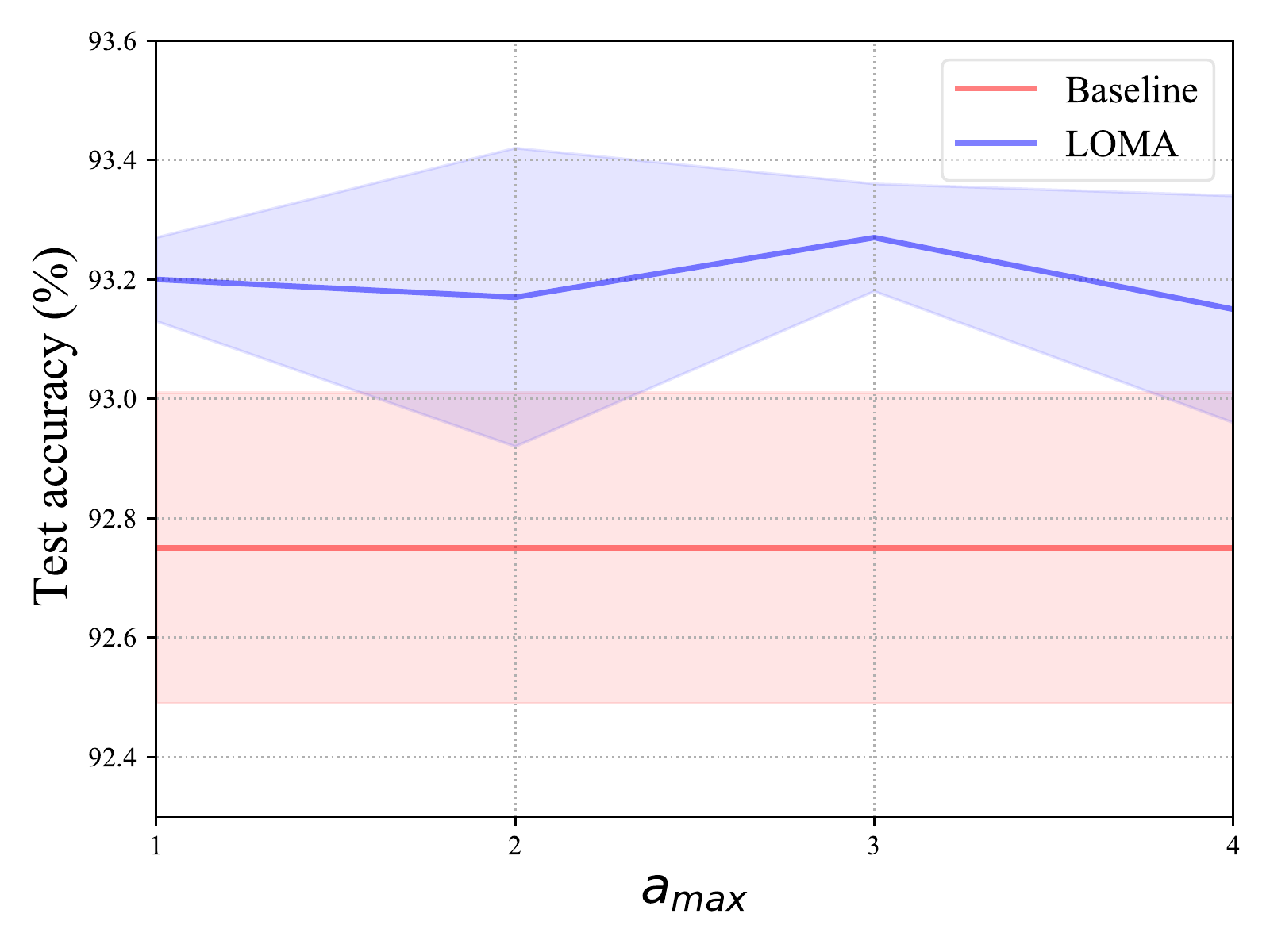}
    \caption{Compression ratio $a_{max}$}
    \label{fig:subfig:c}
  \end{subfigure}
  \caption{Test accuracy (\%) with different $p$, $r_{max}$, and $a_{max}$ on CIFAR-10 based on ResNet-20 using \name (rhombus).}
  \label{fig6}
\end{figure*}

\begin{figure*}[ht]
  \centering
  \begin{subfigure}{0.45\linewidth}
    % \fbox{\rule{0pt}{2in} \rule{.9\linewidth}{0pt}}
    \centering
    \includegraphics[width=0.95\linewidth]{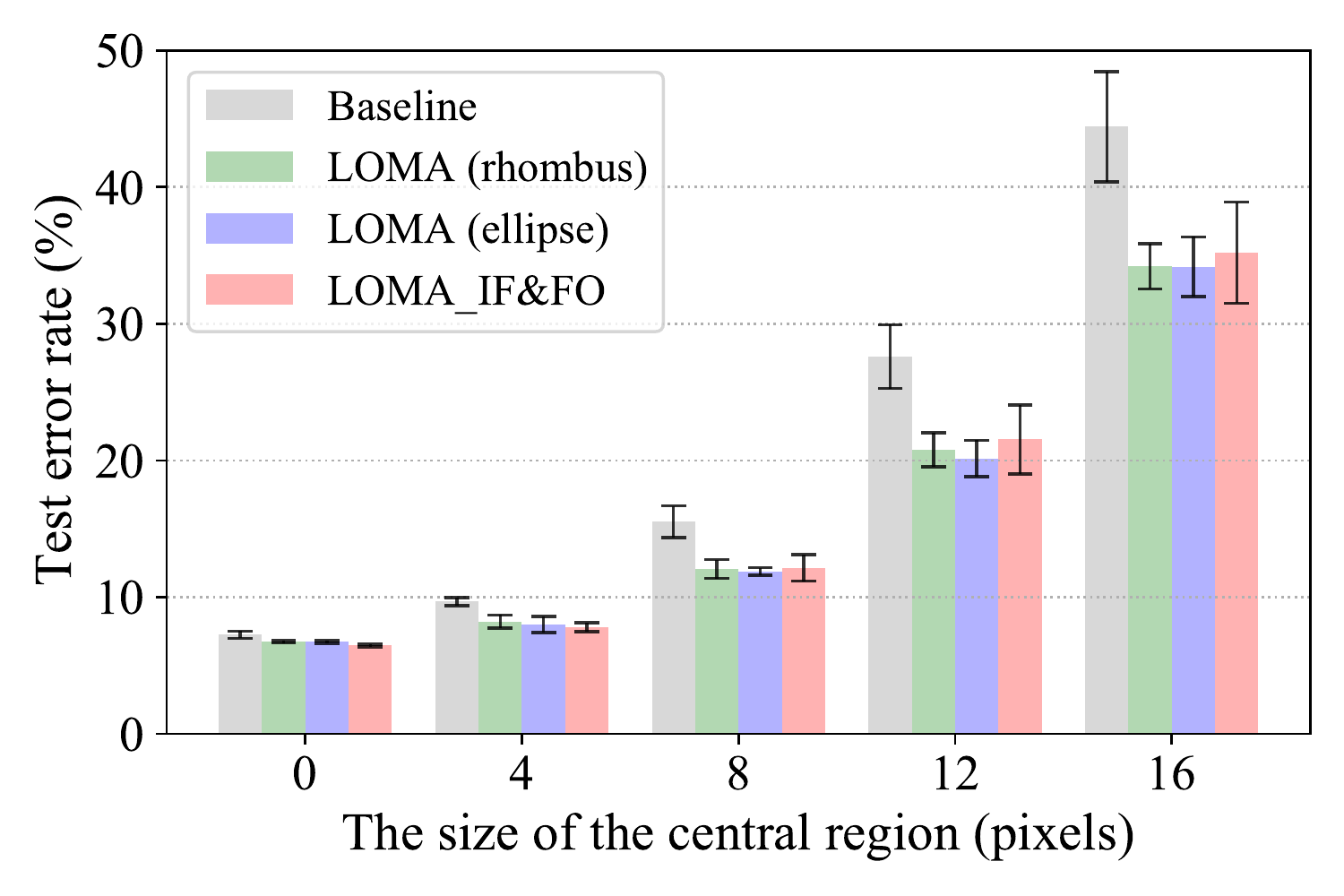}
    \caption{Center occlusion}
    \label{fig7:subfig:a}
  \end{subfigure}
  \hfill
  \begin{subfigure}{0.45\linewidth}
    \centering
    % \fbox{\rule{0pt}{2in} \rule{.9\linewidth}{0pt}}
    \includegraphics[width=0.95\linewidth]{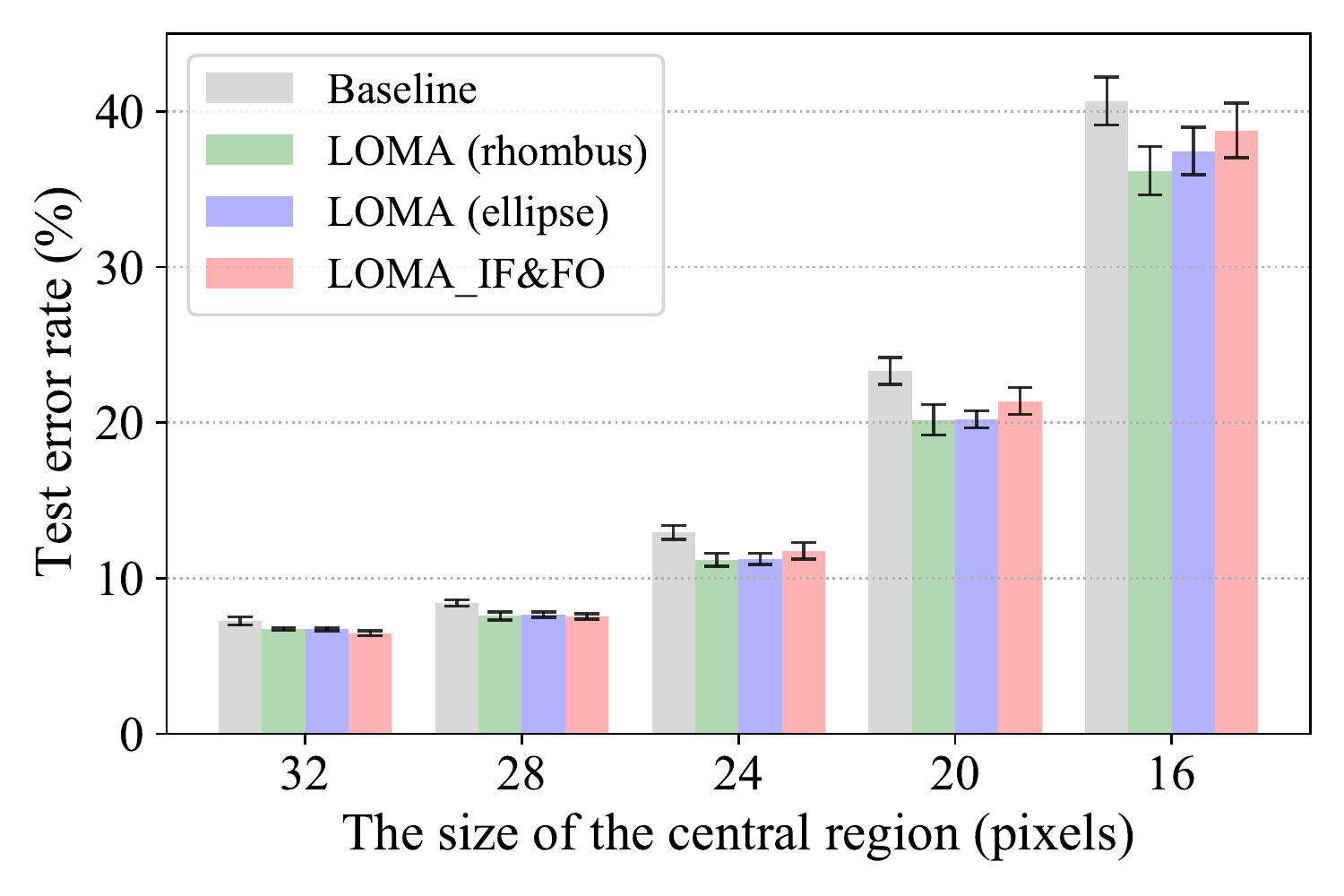}
    \caption{Boundary occlusion}
    \label{fig7:subfig:b}
  \end{subfigure}
  \caption{Robustness to occlusion on CIFAR-10 based on ResNet-20. A lower test error rate is better. }
  \label{fig7}
\end{figure*}

Following the training recipes in CutMix~\cite{cutmix}, we train ResNet-50 with \name and the standard data augmentations for 300 epochs using a weight decay of 1e-4. The batch size is set to 256. The learning rate starts from 0.1 and decays by a factor of 0.1 at epochs 75, 150 and 225, respectively. Since a more complicated dataset calls for an increased probability of augmentation, we set $p=0.8$ for ResNet-50.

%As shown in \cref{table2}, \name and \nameAll improved the accuracy of ResNet-50 by 0.37\% and 0.67\% on a solid baseline, respectively, which indicates that \name and \nameAll can consistently improve the accuracy of the baseline classifier for large-scale datasets. As shown in \cref{fig5}, \name can force the network to focus on a larger range or more important region than the baseline model, which will benefit the network performance. We believe that this occurs because the local magnification augmentations condition the network to model a significantly expanded appearance subspace for an object class without extending into the subspaces of other object classes.

The results on ImageNet are summarized in \cref{table2}. 
To be consistent with the experiments on CIFAR datasets, if not specified, we continue to use \name (rhombus) for the comparison. From the results, we can observe the following: 

\begin{table}[t]
\centering
\resizebox{\columnwidth}{!}{
\begin{tabular}{l|cccc}
\hline
           & Baseline + \name & $l=1$   & $l=2$   & $l=3$   \\ \hline
Top-1 Err. & 17.72 $\pm$ 0.23    & 17.55 $\pm$ 0.12 & \textbf{17.11} $\pm$ 0.23 & 17.91 $\pm$ 0.58 \\ \hline
\end{tabular}
}
\caption{Test errors (\%) of \nameF and FO applied to the $l$-th convolutional block of the network on CIFAR-100 based on WideResNet-28-10.}
\label{table3}
\end{table}

\begin{table}[t]
\centering
\resizebox{\columnwidth}{!}{
\begin{tabular}{l|cccc}
\hline
           & Baseline + \name & $p_f=0.25$ & $p_f=0.5$ & $p_f=0.75$  \\ \hline
Top-1 Err. & 17.72 $\pm$ 0.23    & 17.32 $\pm$ 0.65  & \textbf{17.11} $\pm$ 0.23 &  17.37 $\pm$ 0.10   \\ \hline
\end{tabular}
}
\caption{Test errors (\%) for different probability $p_f$ of \nameF and FO on CIFAR-100 based on WideResNet-28-10.}
\label{table4}
\end{table}

\begin{itemize}
\setlength{\itemsep}{0pt}
\setlength{\parsep}{0pt}
\setlength{\parskip}{0pt}
\item  When only augmenting the input images, \name (rhombus) and \name (ellipse) both boost the performance on the baseline (with standard data augmentation) and show similar or even better results over Cutout and Random Erasing.
\item The results show continuous improvement after superimposing \name, \nameF, and FO sequentially on the baseline model (with standard data augmentation).  \nameIF and \nameAll both outperform the two advanced intensity transformation data augmentation methods, Cutout and Random Erasing. 
 \end{itemize}

Moreover, as shown in \cref{fig5}, \name can force the network to focus on a larger range or more important region than the baseline model, which will benefit the network performance. We believe that this occurs because the local magnification augmentations condition the network to model a significantly expanded appearance subspace for an object class without extending into the subspaces of other object classes.

\subsection{Ablation Study}
\label{Ablation Study}

We explore the impact of hyper-parameters on \name, \nameF and FO in training the CNNs. In order to simplify the experiments, we conduct experiments on CIFAR-10 based on ResNet-20 using \name under different hyper-parameter settings, with the rhombus as the preset deformation shape. As a basic setting, we set $p = 0.5$, $r_{max} = 0.7$, $r_{min} = 0.03$, $a_{max} = 3$, and $a_{min} = 1$ for \name. We experiment with different radius ranges, compression ratio ranges, and probabilities, and then examine their influence on the results. When modifying any one of them, the other two hyper-parameters are fixed. The results are summarized in \cref{fig6}.

% \begin{table}[ht]
% \centering
% \begin{tabular}{l|c}
% \hline
%                       & Top-1 Err. \\ \hline
% Baseline + \name               & 17.72      \\ \hline
% + \nameF &       \\
% + FO     & 17.51      \\ \hline
% + LOMA\_F\&FO               & 17.11$\pm$0.23      \\ \hline

% \end{tabular}
% \caption{Test errors (\%) of adding different components to WideResNet-28-10 on CIFAR-100.}
% \label{table5}
% \end{table}

% \begin{table}[t]
% \centering
% \resizebox{\columnwidth}{!}{
% \begin{tabular}{c|ccc|c}
% \hline
% Method                    & +  LOMA\_I & + LOMA\_F & + FO & Accuracy (\%) \\ \hline
% \multirow{7}{*}{Baseline} &                &           &      &  81.32 $\pm$ 0.20 \\
%                           &  $\checkmark$  &           &      &  82.28 $\pm$ 0.23  \\
%                           &            & $\checkmark$  &      &  82.06 $\pm$ 0.10 \\
%                           &            &           & $\checkmark$ &               \\
%                           & $\checkmark$  &  $\checkmark$  &      & 82.65 $\pm$ 0.30  \\
%                           &            & $\checkmark$ & $\checkmark$    &            \\
%                           &  $\checkmark$  & $\checkmark$  & $\checkmark$ &  82.89 $\pm$ 0.23 \\ \hline
% \end{tabular}
% }
% \caption{Test errors (\%) of adding different components of our method to WideResNet-28-10 on CIFAR-100.}
% \label{table5}
% \end{table}

\begin{table}[t]
\centering
\resizebox{\columnwidth}{!}{
\begin{tabular}{c|ccc|c}
\hline
Method                    & +LOMA\_I                  & +LOMA\_F                  & +FO               & Accuracy (\%) \\ \hline
                                     
\multirow{7}{*}{Baseline} &                           &                           &                      &81.32 $\pm$ 0.20  \\
                          &       $\checkmark$        &                           &                      &82.28 $\pm$ 0.23 \\
                          &                           &      $\checkmark$         &                      &82.06 $\pm$ 0.10               \\
                          &                           &                           &    $\checkmark$      &81.55 $\pm$ 0.41  \\
                          &        $\checkmark$       &      $\checkmark$         &                      &82.65 $\pm$ 0.30                  \\
                          &                           &      $\checkmark$         &    $\checkmark$      &82.08 $\pm$ 0.27                  \\
                          &        $\checkmark$       &      $\checkmark$         &    $\checkmark$      &\textbf{82.89} $\pm$ 0.23    \\ \hline
\end{tabular}
}
\caption{Test accuracy (\%) of adding different components of our method to WideResNet-28-10 on CIFAR-100.}
\label{table5}
\end{table}

\begin{table}[t]
\centering
\begin{tabular}{@{}c|ccc|c@{}}
\hline
Method                    & + RC & + RF & + \name & Accuracy (\%)  \\ \hline
\multirow{8}{*}{Vanilla } &                 &                 &                       & 89.35 $\pm$ 0.16 \\
                          & $\checkmark$    &                 &                       & 94.29 $\pm$ 0.14 \\
                          &                 & $\checkmark$    &                       & 92.50 $\pm$ 0.18 \\
                          &                 &                 & $\checkmark$          & 90.66 $\pm$ 0.36 \\
                          & $\checkmark$    & $\checkmark$    &                       & 95.32 $\pm$ 0.18 \\
                          & $\checkmark$    &                 & $\checkmark$          & 94.99 $\pm$ 0.31 \\
                          &                 & $\checkmark$    & $\checkmark$          & 93.60 $\pm$ 0.19 \\
                          & $\checkmark$    & $\checkmark$    & $\checkmark$          & \textbf{95.82} $\pm$ 0.14 \\ \hline
\end{tabular}
\caption{Test accuracy (\%) with different data augmentation on CIFAR-10 based on ResNet-18. RC: Random cropping, RF: Random flipping.}
\vspace{-1em}
\label{table6}
\end{table}

\begin{table*}[t]
\centering
\resizebox{\textwidth}{!}{
\setlength{\tabcolsep}{0.6mm}{
\begin{tabular}{@{}lc|cccccccccccccccccccc|c@{}}
\toprule
Method   & train set & aero & bike & bird & boat & bottle & bus  & car  & cat  & chair & cow  & table & dog  & horse & mbike & person & plant & sheep & sofa & train & tv   & mAP  \\ \midrule
Faster RCNN & 07        & 75.1 & 82.3 & 71.1 & 55.5 & 59.4   & 78.9 & 85.3 & 80.9 & 56.2  & 76.4 & 65.9  & 79.5 & 81.9  & 76.9  & 83.2   & 50.1  & 73.0  & 68.9 & 78.3  & 71.2 & 72.5 \\
LOMA (rhombus) & 07        & 82.0 & 81.9 & 73.2 & 59.4 & 59.4   & 79.6 & 85.2 & 80.6 & 59.0  & 70.1 & 63.3  & 78.8 & 80.9  & 81.3  & 83.4   & 50.5  & 68.8  & 71.4 & 80.0  & 73.1 & \textbf{73.2} \\ 
LOMA (ellipse) & 07        & 80.7 & 83.5 & 70.0 & 58.3 & 61.0   & 79.5 & 86.0 & 81.6 & 59.1  & 72.6 & 64.8  & 79.4 & 81.5  & 80.7  & 83.2   & 50.7  & 67.8  & 70.8 & 80.5  & 71.9 & \textbf{73.2} \\ \midrule
Faster RCNN & 07+12     & 85.1 & 87.5 & 83.6 & 69.8 & 71.4   & 85.2 & 88.1 & 88.4 & 65.4  & 83.9 & 72.7  & 87.5 & 86.2  & 84.8  & 86.0   & 55.5  & 82.5  & 77.6 & 84.9  & 77.5 & 80.2 \\
LOMA (rhombus) & 07+12     & 84.6 & 86.7 & 83.8 & 69.0 & 70.8   & 84.9 & 88.1 & 88.6 & 65.5  & 84.9 & 73.1  & 87.9 & 85.4  & 85.6  & 85.8   & 58.7  & 83.6  & 79.0 & 85.1  & 81.0 & \textbf{80.6} \\ 
LOMA (ellipse) & 07+12     & 85.6 & 87.2 & 84.6 & 69.0 & 70.5   & 83.6 & 88.2 & 88.5 & 65.3  & 85.2 & 72.3  & 87.7 & 86.9  & 85.7  & 85.9   & 56.5  & 83.2  & 78.5 & 85.0  & 81.0 & 80.5 \\ \bottomrule
\end{tabular}
} }
\caption{Average precision (\%) on VOC 2007 test set based on Faster RCNN.}
\label{table7}
\end{table*}

It is important to note that no matter how the hyper-parameters are set, \name can always improve the model performance over the baseline. These experiments verify that different $r$ and $p$ can bring different effects to CNNs, as expected. Our method is also found to be robust to the compression ratios for the preset deformation shape.

For \nameF and FO, we explore %the layer $l$ of its application in the network
the layer $l$ of feature argumentation to be applied, the probability $p$, and the necessity of having the two components. We experiment on CIFAR-100 based on WideResNet-28-10 using \name in the basic setting and \nameF and FO under different hyper-parameter settings. 
% In the experiments in \cref{table3,table4,table5}, the baseline model refers to only using \name at the input layer without \nameIF.

As shown in \cref{table3,table4}, applying \nameF and FO after the second convolution block works best. We believe this is because if it is too close to the input layer, the effect of moving the element position is very similar to that of \name in the input layer, thus bringing limited improvement. On the other hand, if it is too close to the output layer, it will affect the semantic information too much.

As shown in \cref{table5}, we explore the contribution of individual components to the experimental results. The results show that all the three proposed methods can improve the classification accuracy separately, and data augmentation at the input layer is the most effective. The effect can be further improved when the three methods are combined, exceeding the baseline model by 1.67\%.
% applying \name on the Feature Map alone or FeatureMap Offset can further reduce the test error rate. The best result is achieved when both are used together, which gains 0.61\% improvement over the baseline. This indicates that the two components applied on the feature map are complementary.

%------------------------------------------------------------------------
\subsection{Robustness to Occlusion}
\label{Robustness to occlusion}
We study the robustness of the trained models with \name against occlusion on CIFAR-10 based on trained ResNet-20. We follow CutMix~\cite{cutmix} to select regions of different sizes in the image's central region and fill zeros in or outside the area to generate samples with different levels of occlusion. 

\Cref{fig7} illustrates the results. The models trained with \name or \nameAll can significantly improve robustness compared to the baseline. Interestingly, in the model's training process, the augmented image generated by \name does not directly block a specific part of the image. Thus, the model will not obtain any occluded samples, but when the level of blocking increases, whether by center occlusion or boundary occlusion, the effect of the model trained with our method is still better than that of the baseline. This proves that \name and \nameAll can bring sufficient robustness against occlusion to the model.

\subsection{Complementary to Standard Augmentation}
%\subsection{Complementary to Standard Data Augmentation}
\label{Complementary to standard data augmentation}
We compare our method with random cropping and random flipping, two of the most commonly used and effective geometric transformation methods, and the results on ResNet-18 are shown in \cref{table6}. 
Here ``Vanilla" indicates training without any augmentation.

We notice that although our method does not perform as well as the standard augmentations when applied alone, it is complementary to them. When the three methods are used together, the error rate is 4.18\%, outperforming the vanilla method by a large margin of 6.47\%. %10.65-4.18\%. 
Adding \name upon the current standard data augmentation (random cropping plus random flipping) further improves the accuracy by 0.5\% on an already very high accuracy of 95.32\%. 
The results indicate that the three methods of geometric transformation can be used together as a new standard data augmentation strategy for classification tasks.

\subsection{Object Detection}
\label{Object Detection}

The PASCAL VOC dataset~\cite{pascal} is a popular benchmark for object detection. The VOC 2007 dataset contains about 5,000 images over 20 object categories in the training/validation (trainval) set and 5,000 images in the test set. The VOC 2012 dataset contains about 11,000 images in the training/validation set. During the training, we use either the trainval in VOC 2007 or the union of VOC 2007 and VOC 2012 trainval as the training set. We evaluate our method’s performance on the VOC 2007 test set and use the mean average precision (mAP) metric. Faster RCNN~\cite{ren2015faster} is employed as the detection model for these  experiments, using the settings in the open source object detection toolbox MMDetection~\cite{mmdetection} based on PyTorch~\cite{pytorch}. Since the backbone of the object detection task is already pre-trained, it is not suitable for further augmenting the feature map, so we only consider using \name here. For \name, we randomly select the location of the magnification center within the entire image and set $p=0.25$, $r_{max}=0.5$, and $a_{max}=1$.

The results are summarized in \cref{table7}. When using VOC 2007 trainval for training, our method can achieve 0.7\% improvements compared to the baseline, without needing to learn any extra parameters. When using the union of VOC 2007 and VOC 2012 trainval as the training set, the mAP of the detector trained with \name surpasses the baseline of Faster RCNN by 0.4\%. The experiments show that our proposed method also benefits object detection tasks.

%-------------------------------------------------------------------------
\section{Conclusion}
Different from global geometric transformation and occlusion methods, our model-free and easy-to-implement augmentation, Local Magnification (\name), randomly magnifies a part of the input image to generate more effective data for improving the model's generalization. It can also complement global geometric transformation methods %(random cropping, random flipping) 
as a new standard data augmentation suite for training CNNs on both image classification and object detection tasks. 
The extension of \name and random cropping to feature augmentation can further boost the classification performance, and outperform the existing advanced data argumentation techniques based on intensity transformation.  
Our work shows two possible augmentation directions for model generalization: one is how to do distortion locally to strengthen the performance, the other is how to adapt effective data augmentation methods to feature augmentation. We will continue to investigate the two directions in our future work.
%We hope our study could inspire more future work on data augmentation methods based on local geometric transformations.
%-------------------------------------------------------------------------
%\section*{Acknowledgments}
%This work is supported by National Natural Science Foundation (U22B2017).

%%%%%%%%% REFERENCES
{\small
\bibliographystyle{ieee_fullname}
\bibliography{egbib}

\begin{thebibliography}{10}\itemsep=-1pt

\bibitem{DAGAN}
Antreas Antoniou, Amos Storkey, and Harrison Edwards.
\newblock {Data Augmentation Generative Adversarial Networks}.
\newblock {\em arXiv preprint arXiv:1711.04340}, 2017.

\bibitem{mmdetection}
Kai Chen, Jiaqi Wang, Jiangmiao Pang, Yuhang Cao, Yu Xiong, Xiaoxiao Li,
  Shuyang Sun, Wansen Feng, Ziwei Liu, Jiarui Xu, Zheng Zhang, Dazhi Cheng,
  Chenchen Zhu, Tianheng Cheng, Qijie Zhao, Buyu Li, Xin Lu, Rui Zhu, Yue Wu,
  Jifeng Dai, Jingdong Wang, Jianping Shi, Wanli Ouyang, Chen~Change Loy, and
  Dahua Lin.
\newblock {MMDetection: Open MMLab Detection Toolbox and Benchmark}.
\newblock {\em arXiv preprint arXiv:1906.07155}, 2019.

\bibitem{seg}
Liang{-}Chieh Chen, George Papandreou, Iasonas Kokkinos, Kevin Murphy, and
  Alan~L. Yuille.
\newblock Deeplab: {Semantic Image Segmentation with Deep Convolutional Nets,
  Atrous Convolution, and Fully Connected CRFs.}
\newblock {\em IEEE Transactions on Pattern Analysis and Machine Intelligence},
  40(4):834--848, 2017.

\bibitem{chen2020gridmask}
Pengguang Chen, Shu Liu, Hengshuang Zhao, and Jiaya Jia.
\newblock {Gridmask Data Augmentation}.
\newblock {\em arXiv preprint arXiv:2001.04086}, 2020.

\bibitem{autoaugment}
Ekin~D. Cubuk, Barret Zoph, Dandelion Man{\'{e}}, Vijay Vasudevan, and Quoc~V.
  Le.
\newblock {AutoAugment: Learning Augmentation Strategies From Data}.
\newblock In {\em {IEEE} Conference on Computer Vision and Pattern Recognition,
  {CVPR}}, pages 113--123, 2019.

\bibitem{randaugment}
Ekin~Dogus Cubuk, Barret Zoph, Jon Shlens, and Quoc Le.
\newblock {RandAugment: Practical Automated Data Augmentation with a Reduced
  Search Space}.
\newblock In {\em Advances in Neural Information Processing Systems 33: Annual
  Conference on Neural Information Processing Systems 2020, NeurIPS}, pages
  702--703, 2020.

\bibitem{imagenet}
Jia Deng, Wei Dong, Richard Socher, Li-Jia Li, Kai Li, and Li Fei-Fei.
\newblock {ImageNet: A Large-Scale Hierarchical Image Database}.
\newblock In {\em {IEEE} Conference on Computer Vision and Pattern Recognition,
  {CVPR}}, pages 248--255, 2009.

\bibitem{cutout}
Terrance DeVries and Graham~W Taylor.
\newblock {Improved Regularization of Convolutional Neural Networks with
  Cutout}.
\newblock {\em arXiv preprint arXiv:1708.04552}, 2017.

\bibitem{Vit}
Alexey Dosovitskiy, Lucas Beyer, Alexander Kolesnikov, Dirk Weissenborn,
  Xiaohua Zhai, Thomas Unterthiner, Mostafa Dehghani, Matthias Minderer, Georg
  Heigold, Sylvain Gelly, Jakob Uszkoreit, and Neil Houlsby.
\newblock {An Image is Worth 16x16 Words: Transformers for Image Recognition at
  Scale}.
\newblock In {\em 9th International Conference on Learning Representations,
  {ICLR}}, 2021.

\bibitem{pascal}
Mark Everingham, Luc Van~Gool, Christopher~KI Williams, John Winn, and Andrew
  Zisserman.
\newblock {The Pascal Visual Object Classes (VOC) Challenge}.
\newblock {\em International Journal of Computer Vision}, 88(2):303--338, 2010.

\bibitem{patchup}
Mojtaba Faramarzi, Mohammad Amini, Akilesh Badrinaaraayanan, Vikas Verma, and
  Sarath Chandar.
\newblock {Patchup: A Regularization Technique for Convolutional Neural
  Networks}.
\newblock {\em arXiv preprint arXiv:2006.07794}, 2020.

\bibitem{dropblock}
Golnaz Ghiasi, Tsung-Yi Lin, and Quoc~V Le.
\newblock {Dropblock: A Regularization Method for Convolutional Networks}.
\newblock In {\em Advances in Neural Information Processing Systems 31: Annual
  Conference on Neural Information Processing Systems 2018, NeurIPS}, pages
  10750--10760, 2018.

\bibitem{fastrcnn}
Ross~B. Girshick.
\newblock Fast {R-CNN}.
\newblock In {\em {IEEE} International Conference on Computer Vision, {ICCV}},
  pages 1440--1448, 2015.

\bibitem{pyramidal}
Dongyoon Han, Jiwhan Kim, and Junmo Kim.
\newblock {Deep Pyramidal Residual Networks}.
\newblock In {\em {IEEE} Conference on Computer Vision and Pattern Recognition,
  {CVPR}}, pages 5927--5935, 2017.

\bibitem{hataya2020faster}
Ryuichiro Hataya, Jan Zdenek, Kazuki Yoshizoe, and Hideki Nakayama.
\newblock {Faster Autoaugment: Learning Augmentation Strategies Using
  Backpropagation}.
\newblock In {\em Proceedings of the European Conference on Computer Vision,
  ECCV}, pages 1--16, 2020.

\bibitem{resnet}
Kaiming He, Xiangyu Zhang, Shaoqing Ren, and Jian Sun.
\newblock {Deep Residual Learning for Image Recognition}.
\newblock In {\em {IEEE} Conference on Computer Vision and Pattern Recognition,
  {CVPR}}, pages 770--778, 2016.

\bibitem{huang2018auggan}
Sheng-Wei Huang, Che-Tsung Lin, Shu-Ping Chen, Yen-Yi Wu, Po-Hao Hsu, and
  Shang-Hong Lai.
\newblock {AugGAN: Cross Domain Adaptation with GAN-based Data Augmentation}.
\newblock In {\em Proceedings of the European Conference on Computer Vision,
  ECCV}, pages 718--731, 2018.

\bibitem{puzzlemix}
Jang{-}Hyun Kim, Wonho Choo, and Hyun~Oh Song.
\newblock {Puzzle Mix: Exploiting Saliency and Local Statistics for Optimal
  Mixup}.
\newblock In {\em Proceedings of the 37th International Conference on Machine
  Learning, {ICML}}, pages 5275--5285, 2020.

\bibitem{cifar}
Alex Krizhevsky, Geoffrey Hinton, et~al.
\newblock {Learning Multiple Layers of Features from Tiny Images}.
\newblock {\em {Technical report}}, 2009.

\bibitem{randomcrop}
Alex Krizhevsky, Ilya Sutskever, and Geoffrey~E. Hinton.
\newblock {ImageNet Classification with Deep Convolutional Neural Networks}.
\newblock In {\em Advances in Neural Information Processing Systems 25: Annual
  Conference on Neural Information Processing Systems 2012, NeurIPS}, pages
  1106--1114, 2012.

\bibitem{moex}
Boyi Li, Felix Wu, Ser-Nam Lim, Serge Belongie, and Kilian~Q Weinberger.
\newblock {On Feature Normalization and Data Augmentation}.
\newblock In {\em {IEEE} Conference on Computer Vision and Pattern Recognition,
  {CVPR}}, pages 12383--12392, 2021.

\bibitem{lim2019fast}
Sungbin Lim, Ildoo Kim, Taesup Kim, Chiheon Kim, and Sungwoong Kim.
\newblock {Fast AutoAugment}.
\newblock In {\em Advances in Neural Information Processing Systems 32: Annual
  Conference on Neural Information Processing Systems 2019, NeurIPS}, pages
  6662--6672, 2019.

\bibitem{seg2}
Jonathan Long, Evan Shelhamer, and Trevor Darrell.
\newblock {Fully Convolutional Networks for Semantic Segmentation}.
\newblock In {\em {IEEE} Conference on Computer Vision and Pattern Recognition,
  {CVPR}}, pages 3431--3440, 2015.

\bibitem{pytorch}
Adam Paszke, Sam Gross, Soumith Chintala, Gregory Chanan, Edward Yang, Zachary
  DeVito, Zeming Lin, Alban Desmaison, Luca Antiga, and Adam Lerer.
\newblock {Automatic Differentiation in Pytorch}.
\newblock 2017.

\bibitem{yolo}
Joseph Redmon, Santosh~Kumar Divvala, Ross~B. Girshick, and Ali Farhadi.
\newblock {You Only Look Once: Unified, Real-Time Object Detection}.
\newblock In {\em {IEEE} Conference on Computer Vision and Pattern Recognition,
  {CVPR}}, pages 779--788, 2016.

\bibitem{ren2015faster}
Shaoqing Ren, Kaiming He, Ross Girshick, and Jian Sun.
\newblock {Faster R-CNN: Towards Real-time Object Detection with Region
  Proposal Networks}.
\newblock In {\em Advances in Neural Information Processing Systems 28: Annual
  Conference on Neural Information Processing Systems 2015, NeurIPS}, pages
  91--99, 2015.

\bibitem{ElasticDistortion}
Patrice~Y Simard, David Steinkraus, John~C Platt, et~al.
\newblock {Best Practices for Convolutional Neural Networks Applied to Visual
  Document Analysis.}
\newblock In {\em 7th International Conference on Document Analysis and
  Recognition, ICDAR}, pages 958--963, 2003.

\bibitem{randomflip}
Karen Simonyan and Andrew Zisserman.
\newblock {Very Deep Convolutional Networks for Large-Scale Image Recognition}.
\newblock In {\em 3rd International Conference on Learning Representations,
  {ICLR}}, 2015.

\bibitem{singh2018hide}
Krishna~Kumar Singh and Yong~Jae Lee.
\newblock {Hide-and-seek: Forcing a Network to Be Meticulous for
  Weakly-Supervised Object and Action Localization}.
\newblock In {\em {IEEE} International Conference on Computer Vision, {ICCV}},
  pages 3524--3533, 2017.

\bibitem{dropout}
Nitish Srivastava, Geoffrey Hinton, Alex Krizhevsky, Ilya Sutskever, and Ruslan
  Salakhutdinov.
\newblock {Dropout: A Simple Way to Prevent Neural Networks from Overfitting}.
\newblock {\em The journal of machine learning research}, 15(1):1929--1958,
  2014.

\bibitem{suzuki2022teachaugment}
Teppei Suzuki.
\newblock {TeachAugment: Data Augmentation Optimization Using Teacher
  Knowledge}.
\newblock In {\em {IEEE} Conference on Computer Vision and Pattern Recognition,
  {CVPR}}, pages 10904--10914, 2022.

\bibitem{xu2022masked}
Haohang Xu, Shuangrui Ding, Xiaopeng Zhang, Hongkai Xiong, and Qi Tian.
\newblock {Masked Autoencoders are Robust Data Augmentors}.
\newblock {\em arXiv preprint arXiv:2206.04846}, 2022.

\bibitem{xu2022comprehensive}
Mingle Xu, Sook Yoon, Alvaro Fuentes, and Dong~Sun Park.
\newblock {A Comprehensive Survey of Image Augmentation Techniques for Deep
  Learning}.
\newblock {\em arXiv preprint arXiv:2205.01491}, 2022.

\bibitem{cutmix}
Sangdoo Yun, Dongyoon Han, Sanghyuk Chun, Seong~Joon Oh, Youngjoon Yoo, and
  Junsuk Choe.
\newblock {CutMix: Regularization Strategy to Train Strong Classifiers With
  Localizable Features}.
\newblock In {\em {IEEE} International Conference on Computer Vision, {ICCV}},
  pages 6022--6031, 2019.

\bibitem{wideresnet}
Sergey Zagoruyko and Nikos Komodakis.
\newblock {Wide Residual Networks}.
\newblock In {\em Proceedings of the British Machine Vision Conference,
  {BMVC}}, 2016.

\bibitem{mixup}
Hongyi Zhang, Moustapha Ciss{\'{e}}, Yann~N. Dauphin, and David Lopez{-}Paz.
\newblock {Mixup: Beyond Empirical Risk Minimization}.
\newblock In {\em 6th International Conference on Learning Representations,
  {ICLR}}, 2018.

\bibitem{zhang2019adversarial}
Xinyu Zhang, Qiang Wang, Jian Zhang, and Zhao Zhong.
\newblock {Adversarial AutoAugment}.
\newblock In {\em 8th International Conference on Learning Representations,
  {ICLR}}, 2020.

\bibitem{randomerase}
Zhun Zhong, Liang Zheng, Guoliang Kang, Shaozi Li, and Yi Yang.
\newblock {Random Erasing Data Augmentation}.
\newblock In {\em The Thirty-Fourth {AAAI} Conference on Artificial
  Intelligence, {AAAI}}, pages 13001--13008, 2020.

\bibitem{CAM}
Bolei Zhou, Aditya Khosla, Agata Lapedriza, Aude Oliva, and Antonio Torralba.
\newblock {Learning Deep Features for Discriminative Localization}.
\newblock In {\em {IEEE} Conference on Computer Vision and Pattern Recognition,
  {CVPR}}, pages 2921--2929, 2016.

\end{thebibliography}
}

\end{document}